\documentclass[11pt]{article} 
\usepackage{rldmsubmit,palatino}
\usepackage{graphicx}
\usepackage{wrapfig}

\title{Communications that Emerge through\\
Reinforcement Learning Using a (Recurrent) Neural Network}

\author{
Katsunari ~Shibata\thanks{http://shws.cc.oita-u.ac.jp/\~{}shibata/home.html}\\
Department of Innovative Engineering\\
Oita University\\
700 Dannoharu, Oita 870-1192, JAPAN \\
\texttt{katsunarishibata@gmail.com} \\
}

\begin{document}

\maketitle

\begin{abstract}
Communication is not only an action of choosing a signal,
but needs to consider the context and the sensor signals.
It also needs to decide what information is communicated
and how it is represented in or understood from signals.
Therefore, {\bf communication should be realized comprehensively
together with its purpose and other functions}.

The recent successful results in end-to-end reinforcement learning (RL)
show the importance of comprehensive learning and the usefulness of end-to-end RL for it.
Although little is known, the author's group has shown that a variety of communications
emerge through RL using a (recurrent) neural network (NN).
Here, three of our works are introduced again for the coming leap in this field.

In the 1st one, {\bf negotiation to avoid conflicts} among 4 randomly-picked agents was learned.
Each agent generates a binary signal from the output of its
recurrent NN (RNN), and receives 4 signals from the agents three times.
After learning,  each agent successfully made an appropriate final decision
after negotiation for any combination of 4 agents.
{\bf Differentiation of individuality} among the agents also could be seen.

The 2nd one focused on {\bf discretization of communication signal}.
A sender agent perceives the receiver's location
and generates a continuous signal twice by its RNN.
A receiver agent receives them sequentially,
and moves according to its RNN's output to reach the sender's location.
When noises were added to the signal, it was binarized through learning
and {\bf 2-bit communication} was established.

The 3rd one  focused on {\bf end-to-end comprehensive communication}.
A sender receives {\bf 1,785-pixel real camera image} on which a real robot can be seen,
and sends two sounds whose frequencies are computed by its NN.
A receiver receives them, and
{\bf two motion commands for the robot} are generated by its NN.
After learning, though some preliminary learning was necessary for the sender,
the robot could reach the goal successfully from any initial location. 

\end{abstract}

\keywords{\hspace{-6.5mm}
reinforcement learning (RL), (recurrent) neural network (RNN), negotiation,
signal discretization,\\
\hspace{-6.0mm}grounding communication
}

\acknowledgements{This research has been supported by JSPS KAKENHI Grant Numbers
JP15300064, JP19300070, JP23500245
and many our group members.}

\startmain 

\section{Introduction}
Needless to say that communication is a kind of higher functions for us,
and how language and communication emerge has been discussed actively\cite{DeepCom}.
Especially, establishment of communication protocol, discretization of
communication signals, realization of dynamic communication and also grounding of communication
have been open problems for a long time.
The origin of the research can be seen in \cite{Nakano,Werner,Nego1,Nego2},
and many works have been done as in\cite{Emergence,TalkingHeads}.
However, in most of them, existence of symbols or some special architecture
for communication is presupposed, and the communication is dealt with separately
from the other functions or its original purpose.
However, communication needs not only to choose a signal from possible ones,
but needs to consider the context and the present sensor signals.
It also needs to decide what information should be communicated and how they should be
represented in or understood from signals to achieve some purpose.
It is a really comprehensive function actually.

On the other hand, the impressive results by Google DeepMind\cite{DQN,AlphaGo}
make us notice the importance of end-to-end comprehensive learning and usefulness
of reinforcement learning (RL) using a neural network (NN) for it.
A challenge on this basis can be seen also in the field of communication learning\cite{DeepCom}.

However, actually, although little is known to the public unfortunately,
our group has propounded the importance of end-to-end RL
for the emergence of comprehensive function for more than 20 years\cite{RLDM17},
and has shown already several works also in the emergence of communication.
In this paper, they are introduced at this timing to contribute the coming leap in this research field hereafter.


Before entering the introduction of our works, our basic standpoint or setting for communication learning is explained at first.
There are two or more learning agents each of which has its own (recurrent) NN.
Each communication signal is generated based on the output of an agent's NN,
and becomes the input of one or some other agents' NNs.
The agents learn independently based on reward or punishment without any communication
for learning, that means learning is completely decentralized.
No knowledge about what information should be communicated is given at all.
A reward or punishment is given only at the end of one episode,
and it is not given directly for the communication signal, but for the result
of one or a series of motions each of which is done based on the communication.

The emergence of three types of communication are introduced here.
In all of them, communication protocol was acquired through learning.
In the 1st one, negotiation to avoid conflicts among multiple agents that needs dynamical communication emerged
through learning using a recurrent neural network (RNN)\cite{Negotiation}.
In the 2nd one, communication signal was binarized through learning in a noisy environment\cite{ComBinary}.
In the 3rd one, grounded communication emerged in which a sender generated communication signals
from raw camera pixel image, and a receiver received them and generated motion commands
for a real robot\cite{E-puck}.
The detail of each work can be found in each reference.

\begin{wrapfigure}[18]{r}{70mm}
\vspace*{-\intextsep}
\vspace{-10mm}
\centering
\includegraphics[height=5.4cm]{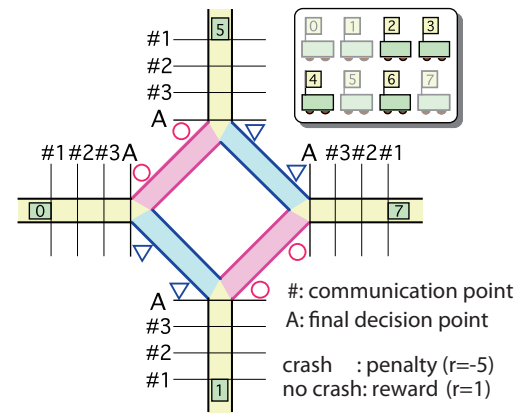}
\caption{Negotiation task. 4 agents are picked up randomly.
They negotiate through 3 chances and decide which path to go finally
to avoid a collision with a neighbor agent. (1999)\cite{Negotiation}}
\label{fig:Nego}
\end{wrapfigure}

\section{Learning of Negotiation to Avoid Conflicts (1999)\cite{Negotiation}}
At first, it is shown that as dynamic communication, negotiation among multiple agents
emerges through learning only from rewards and punishments. 
As shown in Fig. \ref{fig:Nego}, there are 8 agents and 4 of them are picked up randomly
at each episode.
Each of 4 agents is located randomly at one of the entrances of 4 paths.
3 communication chances are given to the agents, and after that
as the final decision, each agent decides which of the two paths, $\bigcirc$ and $\bigtriangledown$,
it will go to.
At each chance, each agent can send a binary signal and receive 4 signals
from the other 3 agents and itself as the inputs for the next time step.
If neighbor agents choose the same path,
they collide and a penalty is given to the both agents.
If an agent can go through the path without collision,
the agent can get a reward, but does not share it with the other agents.
If all the agents choose the path other than their two neighbors
and the same as its opposite,
they can rotate clockwise or anti-clockwise without any collision.
The agent cannot know either the source of each signal or which agents are chosen.

\begin{wrapfigure}[21]{r}{90mm}
\centering
\includegraphics[height=4.5cm]{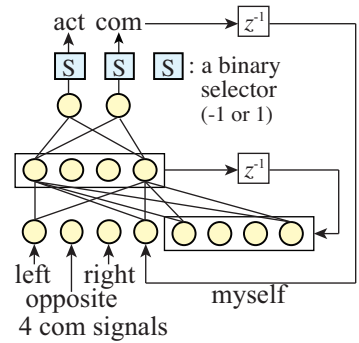}
\caption{A recurrent neural network (RNN) for each agent.\cite{Negotiation}}
\label{fig:NegoNet}
\end{wrapfigure}

Each agent has its own RNN.
As shown in Fig. \ref{fig:NegoNet}, the RNN has two outputs,
one of which is used to decide its communication signal and the other is used to decide its action.
The communication signal or action is stochastically chosen from -1 and 1.
The probability for taking the value 1 is decided by the output of the RNN.
The first communication signal is not depending on the other agents' signals
and always the same after learning.
Here, a primitive type of RL is employed.
When each agent receives a reward, the probabilities of the action and
a series of the communication signals are learned to be large,
and when it receives a penalty, the probabilities are learned to be small.
The RNN is trained by BPTT (Back Propagation Through Time) learning\cite{BP}.
It is not so easy for us to design the communication strategy of each agent,
but all the agents went through a path successfully for any combination of agents after learning.

Two examples of negotiation are shown in Tab. \ref{tab:NegoSample}.
In the case of Table \ref{tab:NegoSample} (a), agents 0, 2, 5, 7 were chosen.
At the 1st communication chance \#1, only the signal from the agent 5 was 1.
At the chance \#2, the agent 0 changed the signal to 1,
and at the chance \#3, all the signals were the same as the chance \#2.
The final decision of each agent was the opposite of the 3rd signal.
By observing many cases, it could be interpreted that the signal 1 means the request to go to
the path $\bigtriangledown$ (action -1).
In this case, if only the agent 0 changed its mind, then no conflict occurred.
The agent 0 seems to have understood the situation from the signals at the communication chance \#1
and changed its decision flexibly at the chance \#2.

\begin{wraptable}[6]{r}{90mm}
\vspace*{-\intextsep}
\centering
\vspace{-23mm}
\caption{Two examples of negotiation after learning.\cite{Negotiation}}
\vspace{1mm}
\includegraphics[height=3.5cm]{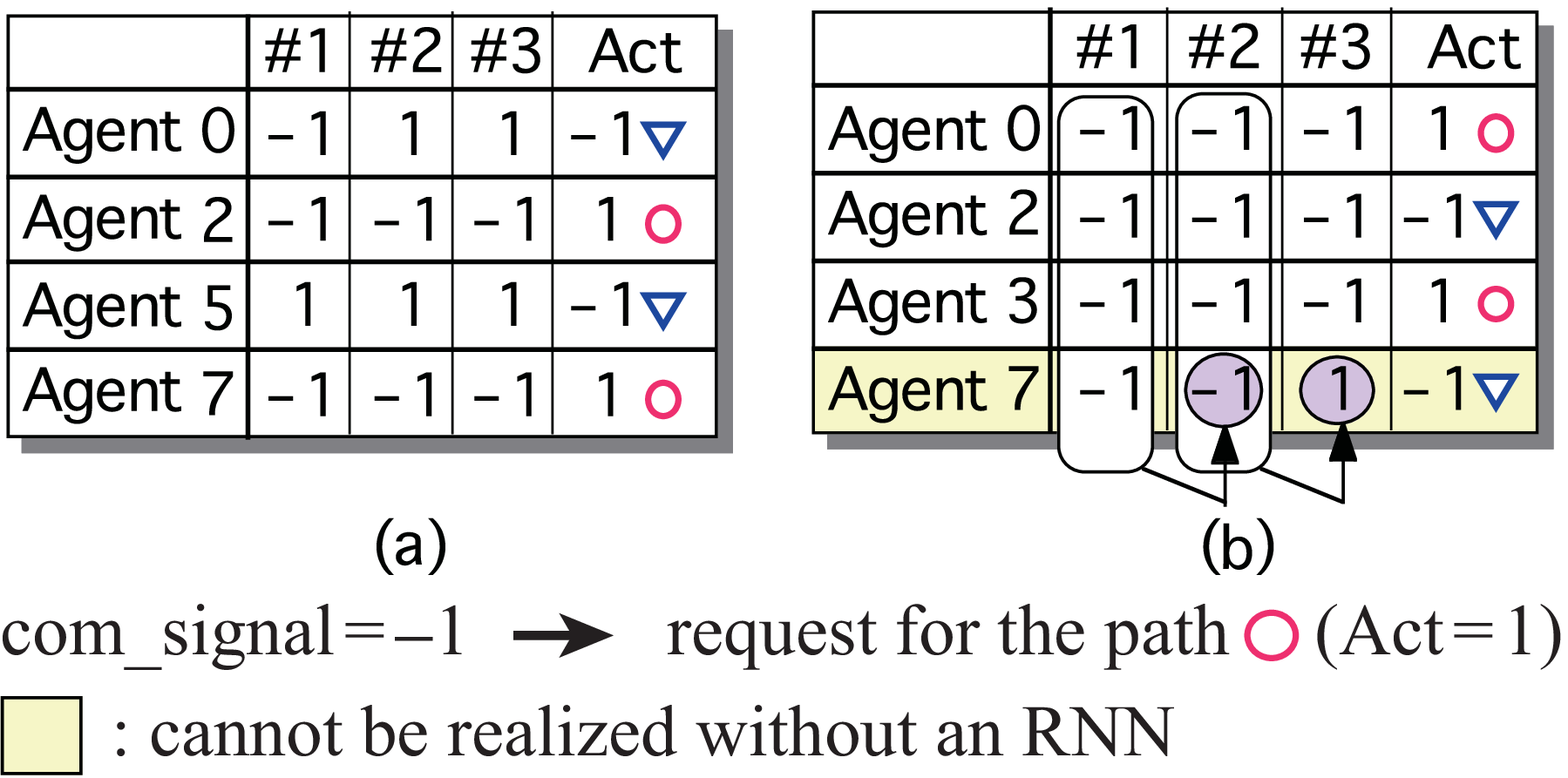}
\label{tab:NegoSample}
\end{wraptable}

In the case of Table \ref{tab:NegoSample} (b), agents 0, 2, 3, 7 were chosen.
The first and second communication signals for 4 agents were the same,
and no one gave up their preferences.
However, at the communication chance \#3, the  agent 7 changed its signal 
even though the inputs of the RNN were completely the same as the chance \#2.
Then,  the  agent 2 that was located at the opposite side of the agent 7
understood the situation and changed the final action,
and then the agent 0 and 3 took the action 1 (path $\bigcirc$) and the others
took the  action -1  (path $\bigtriangledown$).   

The interesting thing is that though the RNN was the same among the agents
(the initial weights were set randomly) and learning was completely independent,
the individuality emerged through learning.
The most stubborn agent hardly gave up its preference,
and the most indecisive agent changed its mind soon
when some conflict occurred.

\section{Emergence of Signal Discretization (2005)\cite{ComBinary}}
We use language or symbols for communication.
The symbols can be considered as the result of
discretization of communication signals.
Therefore, if communication signals are discretized through learning,
it is very interesting from the viewpoint of solving the mystery of symbol emergence.

A simple reason for the emergence  of signal discretization that is thought of easily
is ``the need for correct information transmission even in a noisy environment''.
In \cite{DeepCom}, possibility of the emergence was suggested
from the result of centralized learning in a special structure.
Before that, the author examined discretization of continuous-valued
communication signals from a need through decentralized reinforcement learning (RL)
as follows.
The details can be found in \cite{ComBinary}.

\begin{wrapfigure}[13]{r}{80mm}
\vspace*{-\intextsep}
\centering
\includegraphics[height=2.6cm]{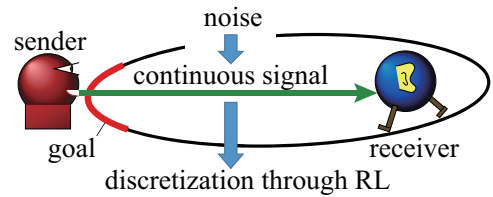}
\caption{A one-way communication task to examine the binarization of continuous-valued 
signals by noise addition through learning.\cite{ComBinary}}
\label{fig:Com}
\end{wrapfigure}
\begin{figure}
\center
\includegraphics[height=5.2cm]{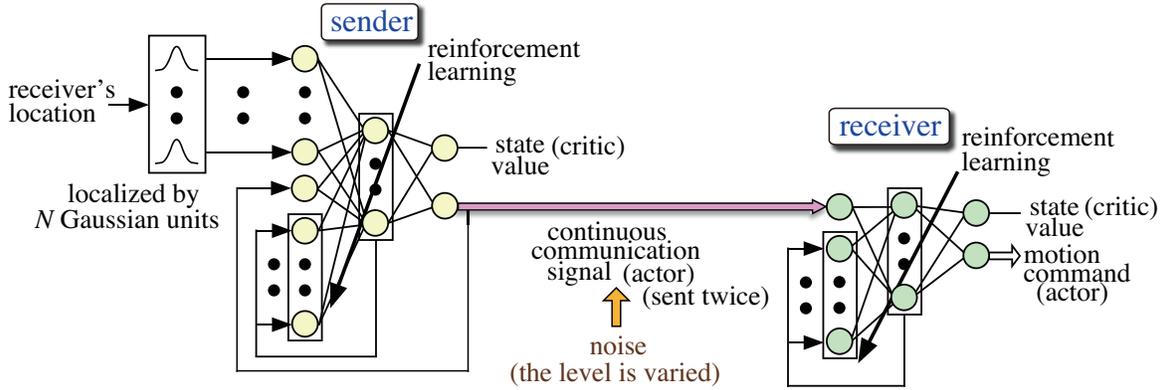}
\caption{Signal flow in the one-way communication task.
Each of the sender and the receiver has its own RNN.
Continuous-valued communication signal that is the actor output of the sender's RNN
is given to the receiver's RNN directly as inputs.
The receiver moves according to its actor output after receiving two signals sequentially
from the sender.
Change in the signal representation by the noise addition
is observed.(2005)\cite{ComBinary}}
\label{fig:ComSys}
\end{figure}

A very simple one-way communication task is assumed.
As shown in Fig. \ref{fig:Com}, there are two agents, which are a sender and a receiver,
and they exist in a one dimensional ring-shape space.
Each of them has their own recurrent neural network (RNN), and learns to communicate.
In each RNN, there are two outputs, actor and critic, as shown in Fig. \ref{fig:ComSys}.
The sender has a sensor to perceive the receiver's location,
but cannot move.
It receives vision-like 30 sensor signals each of which responds the existence of receiver agent
in a local area of the space, and put the signals to its own RNN as inputs directly.
The actor output of the RNN is sent directly as a continuous communication signal
to the receiver.
The receiver receives the signal, puts it to its own RNN as input, and computes its RNN.
At the next time step, the sender computes its RNN, and the new output (signal) is
sent to the receiver again.
The receiver puts the signal to its RNN as input, and computes its RNN again.
After that, the receiver moves according to the actor output of its RNN.
When the receiver reached the goal close to the sender without overshooting,
they can both get a reward, and learns independently.
The task is very easy such that if the receiver takes appropriate motion,
it can reach the goal in one step from any place in the space.
The both RNNs are learned using the training signals generated
based on the popular TD-type RL for actor-critic.
However, the actor output is not used as the probability for a discrete action,
but used directly as a continuous communication signal from the sender
or continuous motion in the receiver.
\begin{wrapfigure}[19]{r}{90mm}
\centering
\includegraphics[height=4.5cm]{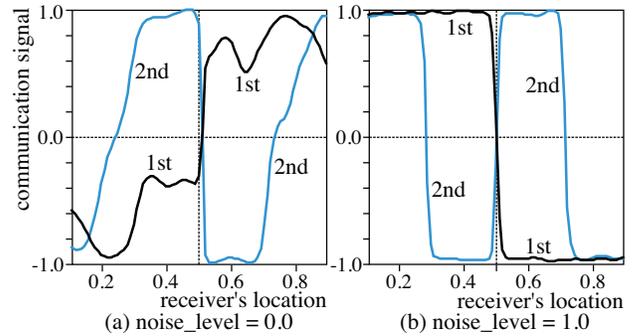}
\caption{Communication signals after learning.
(a) no noise is added and (b) some noises are added to the signals.
The number in the graph indicates the order of the two sequentially transmitted signals.
In (b), 2-bit communication was established without any direction.\cite{ComBinary}}
\label{fig:Signal}
\end{wrapfigure}

After learning, the communication was established, and
the receiver could reach the goal in one step from any location.
Even though what information should be communicated was not given beforehand,
it could be acquired from the necessity only through completely independent learning.
It is also interesting that the receiver's location could be transmitted
using the two signals that were sent sequentially.

Furthermore, noises were added to the signal during the learning phase.
Fig. \ref{fig:Signal} shows the two signals as a function of the receiver's location after learning
for the two cases of no noise and addition of some noise.
The communication signals after learning with no noise seem irregular, but the receiver
could reach the goal in one step.
However, it was difficult to reach the goal in one step in the noisy environment.
On the other hand, when learning in the noisy environment, the sender outputs became almost binary
and 2-bit communication was established.
In this task, to reach the goal in one step, the receiver was required to know its rough location.
Then the sender learned to send which of the four areas in the space the receiver existed
using two bit binary signals, and the receiver learned to interpret the two bit binary signals
and moved appropriately to the goal in one step.
It is surprising that even though the output of the RNN was a continuous value,
and no directions or no options of ``discretization'' were given,
they discovered 2-bit communication by themselves from the purpose of successful goal reaching for the receiver.
The author does not think that the discrete fashion of language or symbols can be explained
only by the above,
but could show the possibility that they emerge through purposive learning
though the dynamics of an RNN should be utilized further.

\section{Learning of Purposive and Grounded Communication (2011)\cite{E-puck}}
Grounding symbol or communication has been a serious problem in AI for a long time\cite{Grounding,Steels}.
Just manipulation of symbols cannot be called communication,
but they should reflect the state of the other agents, objects or environment
that can be perceived through sensors.
Aiming to solve the problem, emergence of lexicon was shown\cite{Nakano,Steels}.
However, they employed special learning for the emergence under a special assumption of
sharing the same observation.
Learning of communication should be realized comprehensively with its purpose and other functions.
Then our group showed that through the general end-to-end RL (actor-critic), 
communication in which signals are generated from a real camera color image as raw sensor signals
emerged for the purpose of leading a real robot to its goal.

\begin{figure}
\center
\includegraphics[height=7.0cm]{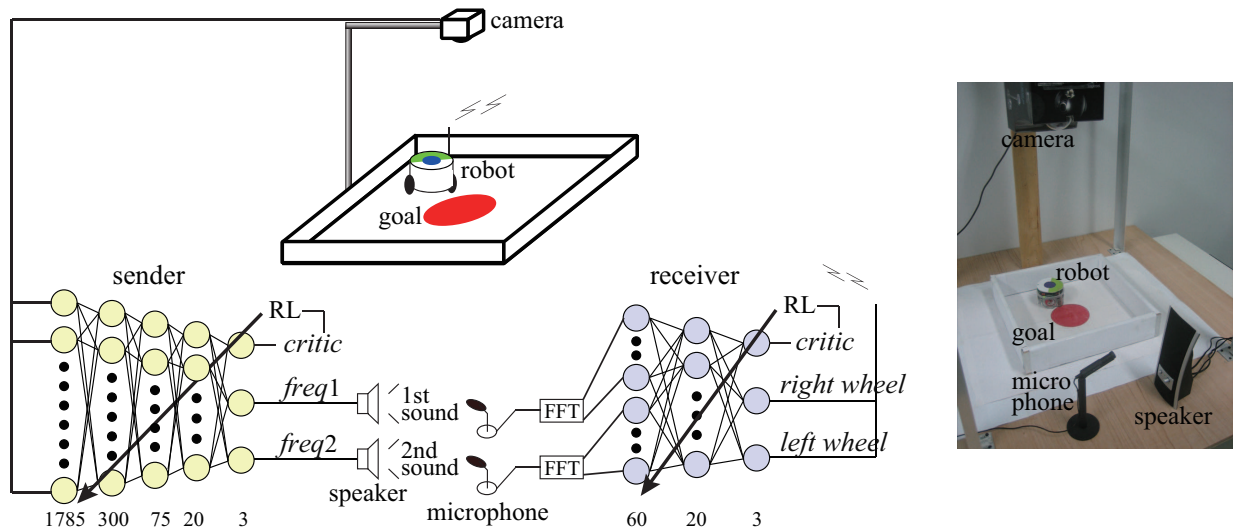}
\caption{Emergence of communication through end-to-end reinforcement learning for the purpose of leading a real robot to the red goal area. The sound signals are generated from raw image signals from a real camera. (2011)\cite{E-puck}}
\label{fig:GroundedCom}
\end{figure}

As shown in Fig. \ref{fig:GroundedCom}, a sender agent receives a color image from the camera
that captures the robot at its center from the above.
The image consists of 26x20 pixels originally, but the resolution of the 5x5 center area is doubled.
Then 1,785 raw camera pixel image signals are the input of the sender's neural network (NN).
Two sounds whose frequencies are decided by the two actor outputs are sent from a real speaker sequentially. 
A receiver receives the two sounds using a real microphone,
and uses them as inputs of its NN after FFT.
The NN outputs the rotation angle for the two wheels of the robot
and they are sent to the robot via bluetooth wireless communication.
They learn independently from the reward that is given when the robot reaches a given goal (red circle area).
However, it was difficult to establish the communication through learning from scratch,
then the sender learned to move the robot directly from the images before this
communication learning.
By using the weights in the hidden neurons after the preliminary learning as initial weights for the communication learning
in the sender agent,
communication was established and the robot could reach the goal after learning.
It is the world's first establishment of communication through end-to-end RL
using a real robot and camera as long as the author knows.\\

\end{document}